# Multimodal trajectory forecasting based on discrete heat map: A Solution to Argoverse Motion Forecasting Competition

Jingni Yuan, Jianyun Xu, Yushi Zhu
HIKVISION Research Institute
Email: lxrbhzj@gmail.com

## 1. Problem Formulation

In Argoverse motion forecasting competition, the task is to predict the probabilistic future trajectory distribution for the interested targets in the traffic scene. We use vectorized lane map and 2 s targets' history trajectories as input. Then the model outputs 6 forecasted trajectories with probability for each target.

## 2. Method

Our solution is modified based on LaneRCNN [2]. The goal candidate sampling and goal candidate scoring method in LaneRCNN are retained. In LaneRCNN, the final prediction result is sampled from the refined goal candidates by non-maximum suppression. While in our method, the goal prediction result of LaneRCNN is used as the intermediate result and the final 6 goals is regressed according to the distribution of the goal candidates with the highest scores. The detailed method is as follows.

### 2.1 Data preprocessing

The target historical trajectory is discretized into motion vectors with an interval of 0.1 s. The vectorized HD maps include the lane topology and lane attributes. We uniformly samples lane nodes on the lane centerlines around the interested target. Then the adjacency matrix of lane nodes is extracted from HD map. Lane node features include the direction, lane segment length, straight / turning, intersection / normal roads and lane speed limit.

Since the vehicles usually obey traffic rules and drive along the lane, we samples goal candidates from the discretized lane nodes for each target. The goal is defined as the end point of 3 s future trajectory. The goal candidates are selected according to the lane structure, topologic connection and vehicle history trajectory. First, the targets' velocity and acceleration are estimated according to the filtered history trajectory. Then, we calculate the correlation between the history trajectory and surrounding lanes, and search all the candidate paths for the interested target. The goal candidate nodes are sampled from the candidate paths according to the current vehicle motion state and kinematic constraints.

### 2.2 Backbone

The backbone of the network is LaneGCN [1], which is also the backbone of LaneRCNN. In LaneGCN, the trajectory feature is extracted by 1D CNN and feature pyramid network. The lane node feature is extracted by two GCN layers. Map to target attention and target to target attention are used to fuse traffic scene features into the target features.

### 2.3 Prediction head

#### 2.3.1 Goal candidate scoring and position refinement

For each target, we concatenate the target feature, the lane node feature and relative position feature corresponding to each goal candidate. For each goal candidate, the goal scoring and refinement model predicts the probability and relative position from the ground truth goal. The goal candidates within 3.0 m from ground truth goal are defined as positive samples, while other goal candidates are defined as negative samples. The refined goal candidates with probabilities constitute a discrete heat map of the predicted goal distribution.

**2.3.2 Goal-based trajectory prediction**

For each target, we sort the goal scores and generate the prediction trajectory for the 70 refined goal candidates with the highest scores (probability). The input of the trajectory prediction network is target feature and the relative position of the refined goal candidate.

**2.3.3 Goal distribution regression**

Based on the discrete goal distribution heat map and the corresponding trajectory for each goal candidate, we extract the global goal distribution feature using PointNet [3]. Each goal candidate is seen as a point in the PointNet. For each target, there are up to 70 points. The input point features include the refined goal candidate position, predicted trajectory and goal candidate score. We use 3 PointNet layers and MLP to generate 6 final goal prediction results based on the input goal distribution. The output includes the 6 goals' position and probability. The trajectory corresponding to each goal is obtained by selecting and refining the nearest trajectory from the previous 70 trajectories.

We use the winner-take-all loss to train the network. The best predicted goal is used to calculate the deviation from the ground truth.

To further improve the goal position and probability prediction accuracy, we use Point Transformer [4] and local attention to refine the goal position and probability. The attention includes self-attention between 6 final goals and attention between 6 final goals and 70 goal candidates.

## 2.4 Tricks

Data augmentation: flip, rotation, scaling transform and dropout of history trajectory points.

Ensemble: We use K-means clustering to fuse the models' results and use weighted average to refine the goal probability. The ensembled models share the same backbone and use different prediction heads.

# 3. Experiments

## 3.1 Ablation study

We compared the metrics of three methods with the same backbone LaneGCN and different prediction heads. The results are shown in Table 1. LaneGCN directly predicts 6 trajectories based on target feature. LaneRCNN uses goal candidate sampling, goal scoring, goal refinement and non-maximum suppression to generate 6 prediction goals. Our method extracts goal candidate distribution features and regresses 6 goals.

Table 1. Ablation study of prediction heads

| Method | Brier-minFDE (Val Set) |
|---|---|
| LaneGCN | 1.7698 |
| LaneRCNN | 1.9518 |
| Our method (without ensemble) | 1.5872 |

As shown in Table 2, model ensemble improves the ranking metric Brier-minFDE by 4.8% on the Argoverse test set. Ensembling prediction results by K-means cluster performs better than match prediction results by bipartite graph matching.

Table 2. Ablation study of ensemble methods

| Ensemble method | Brier-minFDE (Val Set) | Brier-minFDE (Test Set) |
|---|---|---|
| No ensemble | 1.5872 | 1.9118 |
| Bipartite graph matching | 1.5382 | 1.8535 |
| K-means clustering | 1.5158 | 1.8188 |

## 3.2 Results

The results on the Argoverse test set are shown in Table 3.

Table 3. Results on Argoverse test set

| Metrics | Value |
|---|---|
| brier-minFDE (K=6) | 1.8188 |
| minFDE (K=6) | 1.1888 m |
| p-minFDE (K=6) | 2.9324 |
| minADE (K=6) | 0.8179 m |
| p-minADE (K=6) | 2.5615 |
| MR (K=6) | 12.09% |
| DAC (K=6) | 99.09% |
| minFDE (K=1) | 3.5262 m |
| MR (K=1) | 55.29% |
| minADE (K=1) | 1.6286 m |

Goal prediction results in some scenes are visualized in Fig.1. The small grey points are goal candidates sampled from the map. The triangles are refined goal candidates. The six grey circles are the final predicted goals marked with probabilities. The red star is the ground truth goal.

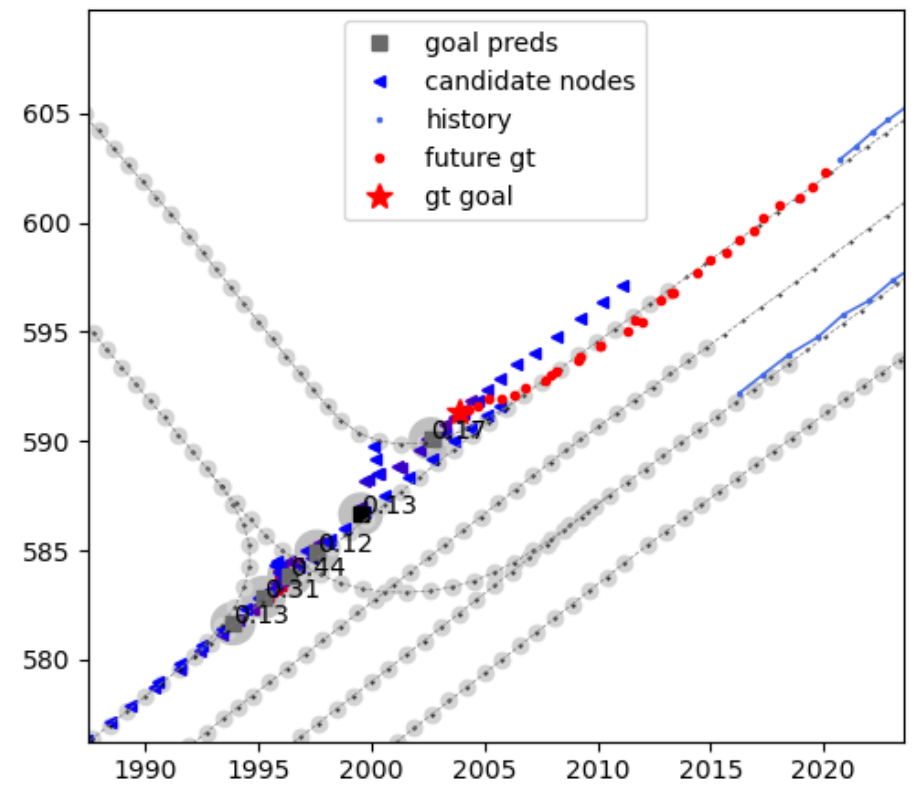

(a) Scene 1

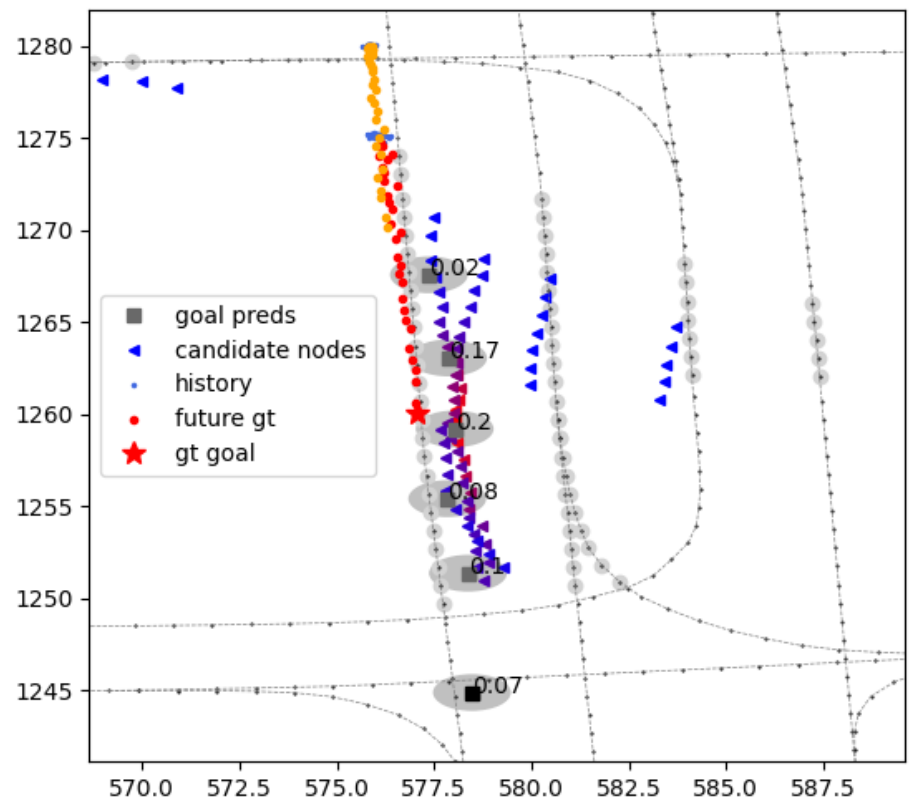

(b) Scene 2

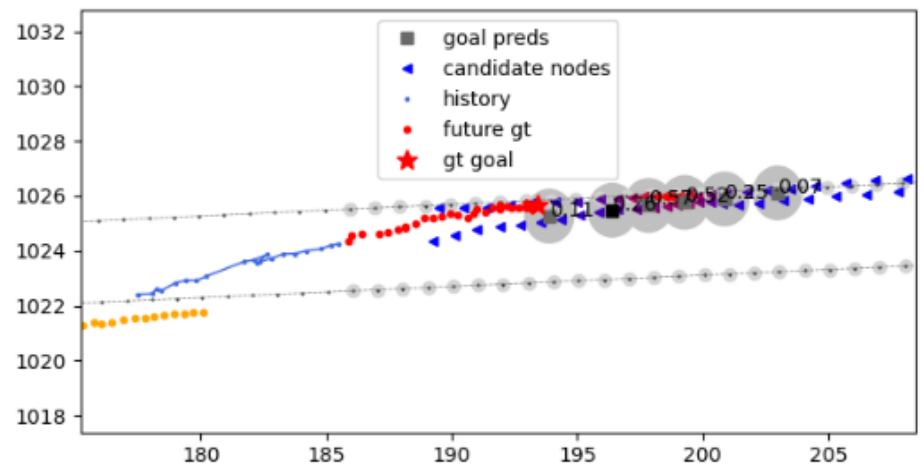

(c) Scene 3

Figure 1. Motion forecasting results